\documentclass[journal,a4paper,twoside]{IEEEtran}

\usepackage [cp1250] {inputenc}
\usepackage{EVjour}

\ifCLASSINFOpdf
   \usepackage[pdftex]{graphicx}
\else 
   \usepackage[dvips]{graphicx}
\fi
\usepackage{epstopdf} 

\usepackage{textcomp} 

\usepackage{amsmath}
\usepackage{amssymb}
\usepackage[noend]{algpseudocode}
\usepackage{algorithm}

\newcommand{\argmax}{\arg\!\max}

\begin{document}

\title{A regularization-based approach for unsupervised image segmentation}

\authors{Aleksandar
Dimitriev${}^{1,\dag}$, Matej Kristan${}^{1,2}$}

\address{$^1$Faculty of Computer and Information Science, University of Ljubljana,\\
Ve\v{c}na pot 113, 1000 Ljubljana, Slovenia \\
$^2$Faculty of Electrical Engineering, University of Ljubljana, \\
Tr\v{z}a\v{s}ka cesta 25, 1000 Ljubljana, Slovenia
\vspace{0.3cm}
\\$^\dag$ E-mail: ad7414@student.uni-lj.si}

\abstract{We propose a novel unsupervised image segmentation algorithm, which aims to segment an image into several coherent parts. It requires no user input, no supervised learning phase and assumes an unknown number of segments. It achieves this by first over-segmenting the image into several hundred superpixels. These are iteratively joined on the basis of a discriminative classifier trained on color and texture information obtained from each superpixel. The output of the classifier is regularized by a Markov random field that lends more influence to neighbouring superpixels that are more similar. In each iteration, similar superpixels fall under the same label, until only a few coherent regions remain in the image. 
The algorithm was tested on a standard evaluation data set, where it performs on par with state-of-the-art algorithms in term of precision and greatly outperforms the state of the art by reducing the oversegmentation of the object of interest.}

\keywords{image segmentation, Markov random field, computer vision}

\received{} 
\review{}    

\markboth{Dimitriev, Aleksandar,  Kristan, Matej}{A regularization-based approach for unsupervised image segmentation}

\maketitle

\IEEEpeerreviewmaketitle


\vspace{2cm}

\section{Introduction}

Image segmentation is a popular branch of computer vision, whose purpose is to partition an image into a number of non-overlapping different segments, 
or labels, that correspond to some meaningful regions of the image.
Depending on the size and number of the resulting regions, the final result
can be called a segmentation or an oversegmentation into superpixels. 

Superpixels are usually very homogeneous in color and texture and respect strong edges. 

Broadly speaking, the image is said to be comprised of superpixels if the number of segments is in the hundreds, though there is no precise definition.
Most methods can produce both coarse and fine segmentations, which can sometimes be controlled by adjusting the method's parameters.
We use the words segment, region, and label interchangeably, since segments 
\begin{figure}[!htb]
    \centering
    \includegraphics[width=0.49\textwidth]{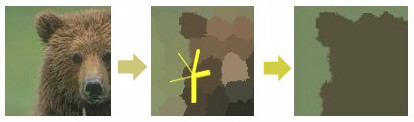}
    \caption{From left to right: input image, an example of MRF pairwise consistency encoded by color similarity, and final segmentation.}
    \label{fig:mrf}
\end{figure}
are defined as all the pixels belonging to the same label.

Our focus in this paper is on non-semantic unsupervised segmentation with an unknown and variable number of segments, i.e. we make no assumptions about the meaning of the regions produced, we do not require any user input and the number of final regions is determined by the output of our algorithm. Therefore, even though a large part of image segmentation concerns supervised segmentation, foreground-background delineation and region labelling, all the work indicated here concerns unsupervised and object-agnostic segmentation unless otherwise noted.

Several approaches to image segmentation have been developed. Among the first were graph-partitioning methods such as 
Shi and Malik's normalized cuts (NCC)~\cite{shi2000normalized} and Felzenszwalb and Huttenlocher's graph cut in nearest-neighbour 
graphs (FH) ~\cite{felzenszwalb2004efficient}, which is frequently used as a preprocessing step in more sophisticated algorithms.
Also well-known is Grabcut~\cite{rother2004grabcut}, though it requires user input and only outputs a foreground/background segmentation.
Another approach is mode-seeking algorithms such as Mean Shift (MS)~\cite{comaniciu2002mean} or Quick Shift~\cite{vedaldi2008quick} 
used in a color space such as RGB or Lab.

Arbelaez et al. ~\cite{arbelaez2011contour} tackle the problem image segmentation through contour detection, 
but the contours are not always valid segmentations because they are not necessarily closed. 
Other unsupervised methods start with a mixture of large number of Gaussians~\cite{yang2008unsupervised} and gradually reduce this number 
by removing degenerated Gaussians and merging those that are closer in feature space than some specified threshold.

Popular methods also include agglomerative clustering, i.e. bottom-up aggregation of either the image's 
pixels~\cite{alpert2012image, joulin2010discriminative, cho2015unsupervised} or superpixels~\cite{mobahi2011segmentation, ren2013image}. 
However, these methods are greedy and suffer from error propagation, where incorrectly merged regions are propagated into subsequent iterations,
although recent work by Wang et al.~\cite{Wang_2015_CVPR} has tried to alleviate such errors by using multiple merge steps.

Although many methods do not use the spatial structure of the image, ignoring this information can sometimes lead to non-smooth segmentations. 
To combat this, many segmentation methods~\cite{wu2007segmentation, zhang2001segmentation, tran2014gaussian, kristan2015fast} 
use MRFs as a way to enforce spatial consistency in neighboring pixels in the image. However, these methods work on the pixel level,
which is becoming increasingly more computationally intensive due to the rise of high-resolution images.
Another benefit of this approach is that by using a superpixel segmentation algorithm that can output a fixed number of superpixels 
all images require roughly the same computational effort regardless of their underlying pixel resolution.

Therefore, in more recent work MRFs have been used on the superpixel level instead as a way to reduce the computational cost and achieve a speed-up~\cite{rantalankila2014generating}. 
Similarly, Fulkerson et al.~\cite{fulkerson2009class} impose a MRF on a preliminary oversegmentation from QS~\cite{vedaldi2008quick}. In contrast to our work, \cite{vedaldi2008quick} require offline pre-training and do not perform segmentation but rather region proposal generation for object detection.
Some methods, such as Li et al.~\cite{li2012segmentation} and Wang et al.~\cite{wang2013graph} use graph partitioning on the superpixel level, 
using methods such as MS~\cite{comaniciu2002mean} and FH~\cite{felzenszwalb2004efficient} as a preliminary step.
However, both methods require that the number of the final segments be known.

\textbf{Our contributions and approach}:
Our work is influenced by several of the aforementioned approaches. We adopt the preprocessing step of oversegmenting the image into several hundred superpixels as in ~\cite{li2012segmentation, wang2013graph, fulkerson2009class}. There exist many methods that are able to provide superpixel intilization such as 
SLIC~\cite{achanta2012slic}, MS~\cite{comaniciu2002mean} and FH~\cite{felzenszwalb2004efficient}. We have found that SLIC works best for our purposes.
From each of these superpixels we extract color and texture features using COLOR-CHILD~\cite{anamandra2014color}, though any descriptor could be used in its place.

The main part of our algorithm are the subsequent iterations. In each iteration we train a discriminative classifier in a 
one-vs-all  fashion, i.e. for each label a binary classifier trained to distinguish between the superpixels belonging to that label (positive examples) and all the others (negative examples).
Similarly to~\cite{fulkerson2009class} we have found that for our purposes Support Vector Machines (SVM)~\cite{cortes1995support} work sufficiently well.
We use these classifiers to re-classify all superpixels to obtain a probability vector of labels for each superpixel.
At the end of each iteration,  we assign each superpixel to the highest probability in the label vector.

Although there are as many labels as there are superpixels in the beginning, which increases the computational cost of each training phase, 
the number of labels quickly declines in subsequent iterations. This is because many labels, especially those at the beginning when each label has a small number
of superpixels, happen not to output the highest probability for any vector and are thus automatically removed from the label pool in all subsequent iterations. 

Lastly, as in~\cite{kristan2015fast, fulkerson2009class}, we adopt the idea of
enforcing spatial consistency of the segmentation using an MRF. Before assigning the superpixels to their new labels, we perform regularization by 
penalizing neighbouring superpixels that are very similar, but have different labels. 
See Figure~\ref{fig:mrf} for an example of superpixels, the pair-wise potentials in the Markov random field, and the final regularized segmentation. 
A more detailed description of each step is subsequently presented, followed by quantitative results and comparison in Section 3.

\section{Methods}

The task of segmenting an image can be formulated as assigning a label $l_i$ to each pixel. The number of labels $K$, however, is unknown a priori. An iterative approach can therefore be applied, that starts with labelling every pixel with its own label and then gradually reduces the number of labels. Our approach is a two-stage approach composed of a pre-segmentation stage and followed by an iterative segmentation stage, which are described next.

\subsection{Pre-segmentation}

Many regions of the image are visually similar and will likely have been assigned the same label in the final segmentation. This means that such neighbouring pixels can be grouped, thus pre-segmenting the image. The image is over-segmented into a few hundred coherent groups of pixels previously defined as superpixels. This can be thought of jump-starting the iterative merging of regions, since merging the pixels in the beginning with a hundred superpixels instead of a million pixels reduces the computational cost of any graph-based methods that operate on the (super) pixel level, at the expense of having slightly less-refined boundaries. The algorithm used in this preliminary step is called Simple Linear Iterative Clustering (SLIC)~\cite{achanta2012slic}. It is a simple sped-up version of k-means that clusters pixels simultaneously in the 5-dimensional CIELAB and coordinate space (L, a, b, x, y). 
Since our method is agnostic of the choice of the superpixel pre-segmentation algorithm, we also experimented with MS~\cite{comaniciu2002mean} and FH~\cite{felzenszwalb2004efficient} as pre-segmentation, and present quantitative differences in Section 3.

\subsection{Super-pixel description and classification}

After oversegmenting the image, the next step is to use a descriptor to obtain discriminative features for each superpixel. The feature descriptor used in our algorithm is COLOR moments augmented Cumulative Histogram-based Image Local Descriptor (COLOR-CHILD)~\cite{anamandra2014color}. The color part contains the first, second, and third image moments of all three color channels, whereas the texture part includes information obtained from first and second-order derivatives. The color and texture features together comprise the $D$ = 57-dimensional descriptor (9 color dimensions and 48-bin quantized histogram) $f_i \in \mathbb{R}^{57 \times 1}$ for the $i^{th}$ superpixel. 

These descriptors can readily be used to learn a classifier for each label. For example, assume we have $M$ superpixels labelled by $K$ labels. A classifier such as a one-versus-all support vector machine (SVM)~\cite{cortes1995support} can be learned for each class from the features extracted from the superpixels labelled by the same label, resulting in $K$ SVMs. The input to the SVM for label $k$ is thus:
\begin{equation}
X = [f_1, ...\;, f_M]^T, \;\;\;\;y = [\delta_{l_1, k}, ...\;, \delta_{l_M, k}]^T,
\end{equation}
where $\delta_{ij}$ is the Kronecker delta function, which is $1$ if $i=j$ and $0$ otherwise. In other words, all the superpixels that are currently labelled as $k$ are the positive examples and all the rest are negative examples, corresponding to the classic one-vs-all fashion for training multi-class SVMs.

Each SVM is calibrated by a Platt calibration such that it outputs a probability of observing the label $l_i$, given its feature descriptor $f_i$. These classifiers are then applied back to each superpixel so that for the next iteration each superpixel is assigned a class label of the SVM with the maximum probability:
$$l_i = \argmax_{j} p(l_j | f_i)$$ 
where $j \in \{1, ...,K\}$.
In this way the superpixels are relabelled. But independent classification of the pixels will likely result in a noisy segmentation and some kind of regularization that penalizes neighbouring superpixels that do not belong to the same class should be enforced.

\subsection{Regularization of segmentation}

To enforce regularization, we apply a Markov Random Field (MRF) on the superpixels. MRFs are first-order graphical models that are commonly used as a way to encode spatial dependencies present between neighbouring pixels in an image. They have found applications in image restoration, stereo vision, and segmentation~\cite{wu2007segmentation,tran2014gaussian,fulkerson2009class,kristan2015fast}. Our approach uses MRFs to take advantage of the structural information present in an image, that would otherwise be unused.  Each superpixel is a variable, with dependencies between superpixels that share a boundary.

The particular type of MRF applied here, and the corresponding procedure of energy minimization, is described in Kristan et al.~\cite{kristan2015fast}. 
Let $\lambda_{ij}$ be the similarity of superpixels $i$ and $j$ defined as
$$ \lambda_{ij} = \frac{1}{Z_i} exp(-||\mathbf{C}_i - \mathbf{C}_j ||^2 )$$
where $\mathbf{C}_i = [r_i, g_i, b_i]^T$ denote the intensity of RGB values of $i$-th superpixel, $||.||$ denotes the usual $l_2$ norm, and $Z_i$ is a normalization constant ensuring the weights sum to $1$, i.e.,
$\sum\nolimits_{j \in N_i} \lambda_{ij} = 1$, where $N_i$ is the neighbourhood of the $i$-th superpixel,
Briefly, the energy function corresponding to the MRF is the following:
\begin{equation}
\label{eq:e}
E \propto \sum\limits_{i=1}^{M} \textrm{log}\; p( l_i | f_i) - \frac{1}{2}(D_{KL}(\pi_i, \pi_{N_i}) + D_{KL}(\textbf{p}_i, \textbf{p}_{N_i})),
\end{equation}
where $\pi_i$ denotes $i$'s prior probability distribution over the labels, $\pi_{N_i}$ is a weighted sum of the priors of $i$'s neighbours $\pi_{N_i} = \sum\nolimits_{j \in N_i, j \neq i} \lambda_{ij}\pi_j$, and $D_{KL}(p, q) = \sum\limits_{x}^{}p(x)log\Big(\frac{p(x)}{q(x)}\Big) $ is the Kullback-Leibler divergence. The variables $\textbf{p}_i$ and $\textbf{p}_{N_i}$ are the corresponding posteriors and the neighbourhood averages, similarly to the priors. The particular formulation of the MRF~\cite{kristan2015fast} treats the priors as well as posteriors as random variables and enforces an MRF on the prior and an MRF on the posterior.

Given the visual likelihoods for all superpixels estimated by the SVMs, i.e., $p( l_i | f_i)$, the posteriors $\mathbf{p}_i$ are computed over all superpixels by minimizing the energy function from (\ref{eq:e}) by the iterative approach from~\cite{kristan2015fast}.


\subsection{Iterative re-labeling and class reduction}

Once the posteriors over superpixels are computed, each superpixel is assigned the label with maximum probability. The classes that do not receive any superpixels in a given iteration are removed from the candidate classes. Then the remaining SVMs are re-learned from the relabelled superpixels. The process of superpixel re-classification, unsupported class removal and SVM-relearning is repeated until convergence. The iterative procedure is summarized in Algorithm~\ref{alg:pseudo}. 

\begin{algorithm}
\caption{Unsupervised MRF-based segmentation}
\label{alg:pseudo}
\begin{algorithmic}[1]

\State \textbf{Input:} Image $I$
\State $(\textbf{f}_i, l_i) \gets$ color-child(spix($I$))     $\forall i$ 
\While{$\exists i: l_i \neq l_{i - 1}$}{}

\For{each label $\textbf{l}$}
	\State train a SVM to compute $p(l_i | f_i)$
	\State minimize the energy function (\ref{eq:e})
\EndFor
\For{each superpixel $i$}
       	\State //Assign each $i$-th superpixel to the MAP estimate of the label
	\State $l_i = \argmax_{j} p(l_j | f_i)$
\EndFor
\EndWhile
\State $\textbf{return}$ labeled image $I$ using final labels $\textbf{l}$

\end{algorithmic}
\end{algorithm}

\section{Results}

We have used the following parameters in our evaluation: we used an $RBF$ kernel for the SVM with $\gamma = 0.001$ and regularization constant $C = 1.0$. These parameters were kept fixed in all experiments.

\begin{figure*}[!htb]
    \centering
    \includegraphics[width=0.95\textwidth]{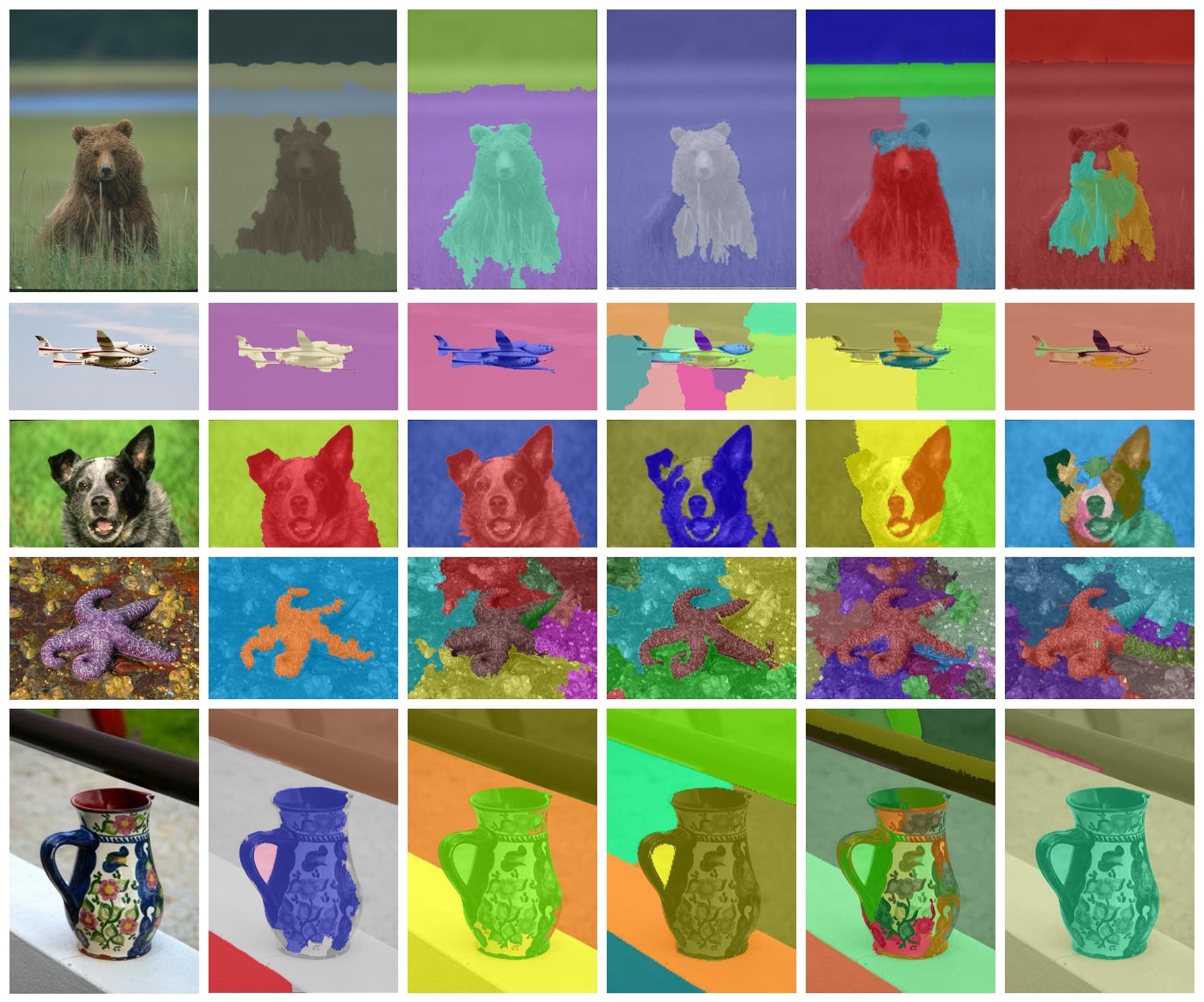}
    \caption{A few images from the dataset and different segmentations. From left to right: Original image, $\text{Our}_{SLIC}$, PBACI, SWA, Normalized cuts, Mean shift.}
    \label{fig:results1}
\end{figure*}

Our algorithm was evaluated on a standard data set~\cite{alpert2012image} consisting of 100 color images which contain a single object of interest that usually occupies a majority of the image (Figure~\ref{fig:results1}). The ground truth consists of three manual foreground-background segmentations provided by three human annotators. The task is to correctly infer the foreground region from the background pixels. The performance is measured by the $F$ measure:
\begin{equation}
F = \frac{2PR}{P + R},
\end{equation}
where $P$ and $R$ denote precision and recall, which measure the fraction of the segment that contains foreground, and the fraction of the foreground that is contained by the segment, respectively. In addition to computing the F measure for each segment and reporting the best value as $F_{single}$, we also computed it for each combination of segments and report the highest value as $F_{multi}$:

\begin{align}
F_{single} &= \max \; \{F_s\} \;\; s \in S \\
F_{multi} &= \max \; \{F_s\} \;\; s \in 2^{|S|} 
\end{align}
where $S$ is the set of segments that comprise the final segmentation, $2^{|X|}$ denotes the power set of $X$, and $F_x$ denotes the F-measure of a segment $x$.

Finally, we assess the fragmentation of each method by counting the number of segments that comprise the combined F-measure as follows:
\begin{equation}
F_\mathrm{frag} = N - 1,
\end{equation}
where $N$ is the number of segments. Lower fragmentation, ideally zero, means that the object is represented by a single segment, whereas high $F_\mathrm{frag}$ implies over-segmentation.

To analyze the results of our method, we compare it to a number of state-of-the-art segmentation algorithms:
\begin{itemize}
  \item Probabilistic Bottom-Up Aggregation and Cue Integration~\cite{alpert2012image}, denoted by PBACI. It gradually merges pixels into successively larger regions by taking into account intensity, geometry, and texture.
  \item Segmentation by weighted aggregation~\cite{sharon2006hierarchy}, denoted by SWA, which determines salient regions in the image and merges them into a hierarchical structure.
  \item Normalized cuts~\cite{malik2001contour}, denoted by N-cuts. It treats the problem of segmentation by computing multiple minimum normalized cuts on a pixel graph.
  \item Contour detection and hierarchical Image Segmentation~\cite{arbelaez2011contour}, denoted by Gpb, which reduces the problem to contour detection and uses spectral clustering to combine local cues into a global framework. 
  \item Mean shift~\cite{comaniciu2002mean}, denoted by MS, a general mode-seeking algorithm on a non-parametric probability distribution, such as the color or intensity distribution.
\end{itemize}

\begin{table}[h]
	\centering
	\begin{tabular}{ | r| c | c | c | }
  		\hline
  		Method  & $F_{single}$ & $F_{multi}$ & $F_\mathrm{frag}$   \\ \hline \hline
  		$\text{Our}_{MS}$ & $0.69 \pm 0.01$ & $\textbf{0.87} \pm 0.01 $ & $ 0.45 \pm 0.03$  \\ 
  		$\text{Our}_{FH}$ & $0.71 \pm 0.01$ & $0.85 \pm 0.01 $ & ${\bf 0.43} \pm 0.03$  \\ 
  		$\text{Our}_{SLIC}$ & $0.72 \pm 0.01$ & $0.84 \pm 0.01 $ & ${\bf 0.40} \pm 0.03$  \\ 
  		PBACI & ${\bf 0.86} \pm 0.01$ & $0.87 \pm 0.02 $ & $1.66 \pm 0.30$  \\ 
  		SWA & $0.76 \pm 0.02$ & $0.86 \pm 0.01 $ & $2.71 \pm 0.33$  \\  
  		N-cuts & $0.72 \pm 0.02$ & $0.84 \pm 0.01 $ & $2.12 \pm 0.17$  \\ 
  		Gpb & $0.54 \pm 0.01$ & ${\bf 0.88} \pm 0.02 $ & $7.20 \pm 0.68$  \\ 
  		MS & $0.57 \pm 0.02$ & ${\bf 0.88 } \pm 0.01 $ & $11.08 \pm 0.96$  \\ \hline
	\end{tabular}
	\smallskip
	\caption[Table caption text]{Results of single and multi-segment coverage on the dataset ($95\%$ confidence). }
	\label{tab:results}
\end{table}

The results are given in Table~\ref{tab:results}, which shows the average scores for all images in the data set. Since we experimented with different preprocessing superpixel segmentations, we denote using $MS$, $FH$ and $SLIC$ with $\text{Our}_{MS}$, $\text{Our}_{FH}$ and $\text{Our}_{SLIC}$, respectively. The best results were achieved using $SLIC$, which had the lowest fragmentation and highest $F$ measure for a single segment. Note that the all variants of our approach deliver the lowest fragmentation error. This means that our iterative approach consistently segments out the objects well, regardless of the initial segmentation process.

Our algorithm delivers highly competitive results in both variants of the $F$ measure. The advantage of our approach is very apparent in fragmentation, where it significantly outperforms the state of the art, which means that it correctly identifies the object with an average of $1.4$ segments regardless of the preprocessing step, whereas all other methods over-segment it. 

It should be noted that there is an inverse relationship between the F measure, specifically $F_{multi}$, and the fragmentation. If a method has high fragmentation, meaning the foreground object is made up of several segments, it is natural to assume that they cover it better than a method that only produces one segment, but the ground truth has only one segment, which should be preferred. Therefore the advantage of our method is correctly delineating the object in the image as being comprised of a single segment. This is because similar superpixels are identified as having the same label early in the iterative process and we are only left with a few segments. 

Note that our method can artificially be made to favor improved multi-segment coverage at the cost of reduced one-segment coverage by increasing the $F_{multi}$ which results in increase of $F_{frag}$. This can be achieved by forcing the SVMs to specialize to the superpixels belonging to their segment, which results in reduced merging of segments. For SVMs this can be achieved by increasing the $\gamma$ parameter, which enhances non-linearity and increases the specialization. A few examples of the (non-overspecialized) segmentation produced by our algorithm are shown in Figure~\ref{fig:results1}.

\section{Conclusion}

An unsupervised iterative segmentation algorithm was proposed. The results show that the algorithm is comparable to the state-of-the-art in precision and recall, and also outperforms the state-of-the-art by more often correctly identifying the segments belonging to a single object. 

Future work will involve considering other classifiers that allow efficient learning and classification of high-dimensional features. The current segment labelling is binary (hard assignment), so using soft-labelling, where a segment would belong to different labels simultaneously, could also be explored. The pairwise MRF energy term, i.e. the edge weight between neighhbouring superpixels depends on color similarity, but could also be extended to texture. 
We will also consider a hierarchical approach in which the segmentation presented in this work acts as a prior on pixel-level segmentation, which is expected to further improve the segmentation quality, by having more refined segment borders. Lastly, saliency detection, the task of determining the important regions of an image, could benefit from our approach as a preliminary step.

\bibliographystyle{IEEEtran}
\bibliography{bibliography}



\begin{IEEEbiographynophoto}{Aleksandar Dimitriev}
received his B.Sc. degree from the Faculty of Computer and Information Science, University of Ljubljana, where he is currently a M.Sc. student. His research interests include probabilistic machine learning and Bayesian statistics, and their applications to computer vision, bioinformatics and other fields of computer science.
\end{IEEEbiographynophoto}

\begin{IEEEbiographynophoto}{Matej Kristan}
Matej Kristan received his Ph.D from the Faculty of Electrical Engineering, University of Ljubljana in 2008. He is an Assistant Professor at the ViCoS Laboratory at the Faculty of Computer and Information Science and at the Faculty of Electrical Engineering, University of Ljubljana. His research interests include probabilistic methods for computer vision with focus on visual tracking, dynamic models, online learning, object detection and vision for mobile robotics.
\end{IEEEbiographynophoto}

\vfill

\label{finish}

\end{document}